\documentclass[letterpaper, 10 pt, conference]{ieeeconf}  

\IEEEoverridecommandlockouts                              

\def\eg{\emph{e.g.,}\xspace} 
\def\ie{\emph{i.e.,}\xspace}

\newcommand{\eqspacing}{\hspace{0.1cm}}

\newcommand{\figref}[1]{Fig.~\ref{#1}\xspace}
\newcommand{\secref}[1]{Section~\ref{#1}\xspace}

\newcommand{\degrees}{^{\circ}\xspace}

\newcommand{\odem}{2.5D orthophoto\xspace}

\newcommand{\numscan}{N\xspace}
\newcommand{\numtraversable}{M\xspace}

\newcommand{\particle}{y^i_t\xspace}
\newcommand{\importanceweight}{w^i_t\xspace}

\newcommand{\descriptorscores}{\textbf{S}\xspace}

\newcommand{\uavdesc}{\mathbf{d}_{uav}(p)\xspace}
\newcommand{\ugvdesc}{\mathbf{d}_{ugv}\xspace}

\newcommand{\lidarfeat}{\mathbf{f}_{lidar}\xspace}
\newcommand{\semanticfeat}{\mathbf{f}_{semantic}\xspace}
\newcommand{\lidarfeati}{\mathbf{f}_{lidar,i}\xspace}
\newcommand{\semanticfeati}{\mathbf{f}_{semantic,i}\xspace}
\newcommand{\diffth}{d_{th}\xspace}

\newcommand{\aerialdescriptors}{\mathbf{D}_{uav}\xspace}

\newcommand{\aerialcoord}{p\xspace}

\newcommand{\rangeonly}{\textsc{range}\xspace}
\newcommand{\rangesemantic}{\textsc{range-semantic}\xspace}

\newcommand{\rangesemantictable}{\begin{tabular}[x]{@{}c@{}}\textsc{range-} \\ \textsc{semantic} \end{tabular}\xspace}

\newcommand{\rangeonlyfull}{\textsc{range-full}\xspace}
\newcommand{\rangesemanticfull}{\textsc{range-semantic-full}\xspace}
\newcommand{\rangeonlyfulltable}{\textsc{\begin{tabular}[x]{@{}c@{}}\textsc{range-} \\ \textsc{full}\end{tabular}}\xspace}
\newcommand{\rangesemanticfulltable}{\textsc{\begin{tabular}[x]{@{}c@{}}\textsc{range-} \\ \textsc{semantic-full}\end{tabular}}\xspace}

\newcommand{\tableScale}{0.9}

\usepackage{graphics} 
\usepackage{epsfig} 
\usepackage{mathptmx} 
\usepackage{times} 
\usepackage{amsmath} 
\usepackage{amssymb}  
\usepackage{graphicx}
\usepackage[font=small]{caption}
\usepackage{subcaption}
\usepackage{booktabs}
\usepackage{xspace} 
\usepackage{paralist}
\usepackage{float}
\usepackage{color}
\usepackage{algorithm}
\usepackage{algorithmic}
\usepackage{flexisym}
\usepackage[left=0.75in,right=0.75in,bottom=0.785in,top=0.75in]{geometry}

\newcommand{\gordonra}{\textcolor{black}}
\newcommand{\gordonrb}{\textcolor{black}}
\newcommand{\gordonrc}{\textcolor{black}}

\title{\LARGE \bf
Semantics for UGV Registration in GPS-denied Environments
}

\author{Gordon Christie$^{1}$ \and Garrett Warnell$^{2}$ \and Kevin Kochersberger$^{3}$
\thanks{$^{1}$Bradley Department of Electrical and Computer Engineering, Virginia Tech, Blacksburg, VA 24061, USA,
        {\tt\small gordonac@vt.edu}}%
\thanks{$^{2}$Computational and Information Sciences Directorate, U.S. Army Research Laboratory, Adelphi, MD 20783, USA,
		{\tt\small garrett.a.warnell.civ@mail.mil}}
\thanks{$^{3}$Department of Mechanical Engineering, Virginia Tech, Blacksburg, VA 24061, USA,
        {\tt\small kbk@vt.edu}}
}

\begin{document}

\maketitle
\thispagestyle{empty}
\pagestyle{empty}


\begin{abstract}
Localization in a global map is critical to success in many autonomous robot missions. This is particularly challenging for multi-robot operations in unknown and adverse environments.  Here, we are concerned with providing a small unmanned ground vehicle (UGV) the ability to localize itself within a 2.5D aerial map generated from imagery captured by a low-flying unmanned aerial vehicle (UAV). We consider the scenario where GPS is unavailable and appearance-based scene changes may have occurred between the UAV's flight and the start of the UGV's mission.  We present a GPS-free solution to this localization problem that is robust to appearance shifts by exploiting high-level, semantic representations of image and depth data.  Using data gathered at an urban test site, we empirically demonstrate that our technique yields results within five meters of a GPS-based approach.
\end{abstract}

\section{Introduction}
\label{sec:introduction}

UAV-UGV collaboration in GPS-denied environments is hard.
Shared instructions often require the coordinate systems of the UAV and UGV to be registered with one another. For example, when a mission control system uses an overhead map to plan paths for a UGV, it becomes critical for the UGV to register its position in this map. For time-sensitive \gordonrb{missions, this must be done quickly,} where visiting several areas before estimating the location may not be possible. GPS can be used to \gordonrb{help} perform registration, but \gordonrb{it} is not always available.

\begin{figure}
\centering
\includegraphics[width=1\columnwidth]{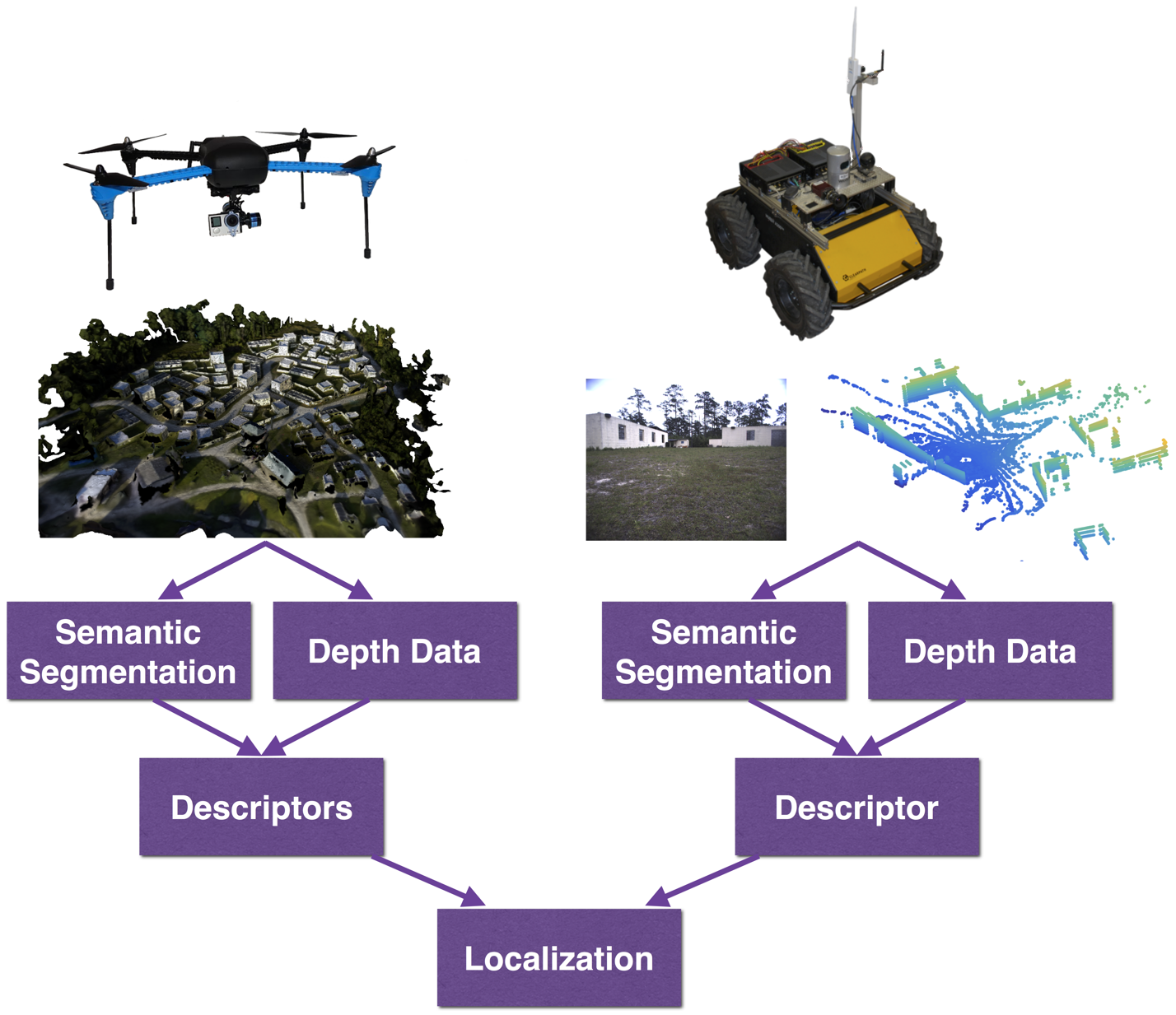}
\caption{An overview of our approach. A UAV captures overhead imagery of a scene to generate a 2.5D orthophoto. Using semantics and depth information, descriptors are created for every traversable pixel in the aerial map. The UGV captures imagery and laser scans. Semantic segmentations and range data are then used to create a descriptor for the UGV data. Descriptor similarities are used to score each traversable pixel in the aerial map, after which a particle filter is used to reason about the location of the UGV.}
\label{fig:teaser}
\end{figure}

Our scenario of interest is one in which multi-robot teams must operate in unknown and adverse environments, and so we consider the problem of localizing a UGV in a map generated by a UAV when GPS information is unavailable. Accurate registration is much more difficult without GPS. The UGV must use data collected from a different perspective \gordonrb{than that of the UAV}, and the scene itself may have changed since the overhead map was generated. In addition, \gordonrb{registration techniques often} rely on data with noisy measurements, imperfect machine learning models, and small errors in sensor calibration.

We perform tests in a challenging urban environment with no \emph{a priori} information (\eg road networks) about the scene. We focus on the case where a UAV flight takes place before the UGV mission. Clearly, the scene may change between the end of the flight and the start of the UGV mission. Structural scene changes (\eg object moves to another location) are one type of change that may occur. The approach we present is robust to \gordonrb{these types of} small scene changes. This is helpful not only for the small changes that may occur, but also for the inherent perspective problem of the task where structures simply look different from the air than the ground. For example, a laser scan from the UGV may not see high enough to observe any structure above a large opening in a building. Therefore, points returned within the range of angles that capture this opening will not match well to the corresponding parts of the aerial map that observe a roof. 

\gordonrb{
Appearance-based scene changes (\eg trees losing leaves) are another concern. 
Matching color information directly has the potential to help immensely, but will likely fail in the presence of these appearance-based changes. We therefore do not use such low-level representations. 
We use semantic segmentations of the aerial and ground data to classify points with category labels (\eg grass). This creates a high-level representation of the scene's appearance, 
where pixels and 3D points are now represented by semantic categories instead of raw color values. This makes our approach robust to appearance-based scene changes. 
}

We propose a GPS-free solution that requires only image, LiDAR, and vehicle odometry data. The contributions are:
\begin{compactenum}
\item A multi-robot system capable of autonomously understanding a scene in GPS-denied environments via joint semantic reasoning about the scene from appearance and depth data.
\item A UGV localization algorithm shown to localize a UGV in an urban environment with an average \gordonrc{difference to GPS} under 5m, where the algorithm is robust to appearance-based scene changes, small structural scene changes, and \gordonrc{occasional ambiguous regions}.
\end{compactenum} 


%


\section{Related Work}
\label{sec:related_work}

The problem of localizing image and LiDAR data in
overhead maps has been the focus of several previous works, \gordonrb{including those that consider} (1) global location estimation of images, (2) localization of image data in an overhead map of a local area, and (3) \gordonrc{our problem; localizing UGV data in a local overhead map with high-level scene representations.}

\noindent \textbf{Global Localization of Images.} The problem of 
\gordonrb{directly estimating geo-location from images}
has been studied in several works~\cite{hauff2012geo,serdyukov2009placing,van2010combining}. 
In \cite{workman2015wide}, deep convolutional neural networks \gordonrb{(CNNs)} are used to perform geolocalization of ground-level query images by matching to georeferenced aerial images. 
\cite{lee2015predicting} use CNNs to recognize geo-informative attributes (\eg population density).
More recently, \cite{weyand2016planet} used CNNs to perform global localization of an image, where they extend their model to incorporate an LSTM that reasons about temporal coherence to localize an entire photo album.
For our task, we are focused on a small area of interest represented by a \odem generated by low-flying UAV imagery. We estimate a precise location of the UGV, where we leverage 2D, 3D, and semantic information about the scene.
With much more data, we believe these other approaches that recognize general areas could be integrated with our approach. A two-stage approach to localize a UGV precisely anywhere on the globe would then be possible.

\noindent \textbf{Local Localization of Images.} An approach to register video with structure from
motion point clouds with temporal contstraints was developed by \cite{kroeger2014video}.
Contrary to our work, they use low-level representations of scene appearance, which we argue will fail in many scenarios. 
In \cite{david2011orientation}, a vision-only approach was used to localize a UGV in a satellite image with manually-defined edges of buildings. Descriptor matching was performed, where descriptors describing a 360$^\circ$ view were calculated for pixels in the satellite image. Similar descriptors were calculated from the ground by identifying building edges in omnidirectional images taken from the on-board camera. Their work inspired a similar descriptor-based approach used in our work. However, our approach includes depth, semantic, and temporal information to perform localization. We also localize the UGV in an aerial map generated by UAV imagery with imperfect depth data, where we automatically label obstacles and semantic categories without human supervision.

\noindent \textbf{Our problem.} In \cite{brubaker2013lost,brubaker2016map}, road networks and visual odometry are used to
perform localization with distributed computation for real-time performance. 
\gordonrb{Our technique does not rely on \emph{a priori} road network information.}
We believe the most similar work to ours is \cite{viswanathan2014vision}. With a similar philosophy, they perform vision-based robot localization in a satellite image across seasons with segmentation outputs. 
However, they do not perform localization in \gordonrb{a complicated urban environment} and do not \gordonrb{exploit} elevation data available from the satellite view to asssist in localization.


\section{Approach}
\label{sec:approach}

We propose an \gordonrc{approach} that integrates range, semantic, and trajectory information to localize a UGV in an aerial map. 
\gordonrb{We pose the problem as one of} finding a mapping between the UGV's trajectory, generated without GPS, to coordinates in the \odem. In Section~\ref{sec:experiments}, we describe how we generate the semantic segmentations used. 

\subsection{Descriptors and Scoring}
\label{sec:descriptors_scoring}

To localize the UGV in the aerial map, we score the similarity \gordonrb{between descriptors generated} from UGV data \gordonrb{and similar descriptors generated with the UAV data}. Our proposed descriptors include range and semantic information to describe each pixel in the \odem, and each local image and laser scan from the UGV. 
We use the same process to generate both the aerial and ground descriptors. We define $\numscan$ scan lines in a 360$\degrees$ view, where the angle between subsequent scan lines is $\alpha = \frac{360\degrees}{\numscan}$. In our experiments, we use $\numscan = 60$, $\alpha = 6\degrees$. We search along each scan line until an obstacle is detected or the max distance (40m) is reached. If an obstacle is detected, then the appropriate element of the descriptor is set to the distance of the obstacle from the current pixel position in the \odem or the origin of the laser scan from the UGV. If no obstacle is detected, then an `invalid' label is assigned to the descriptor element. 

\begin{figure}
	\centering
	\includegraphics[width=0.9\columnwidth]{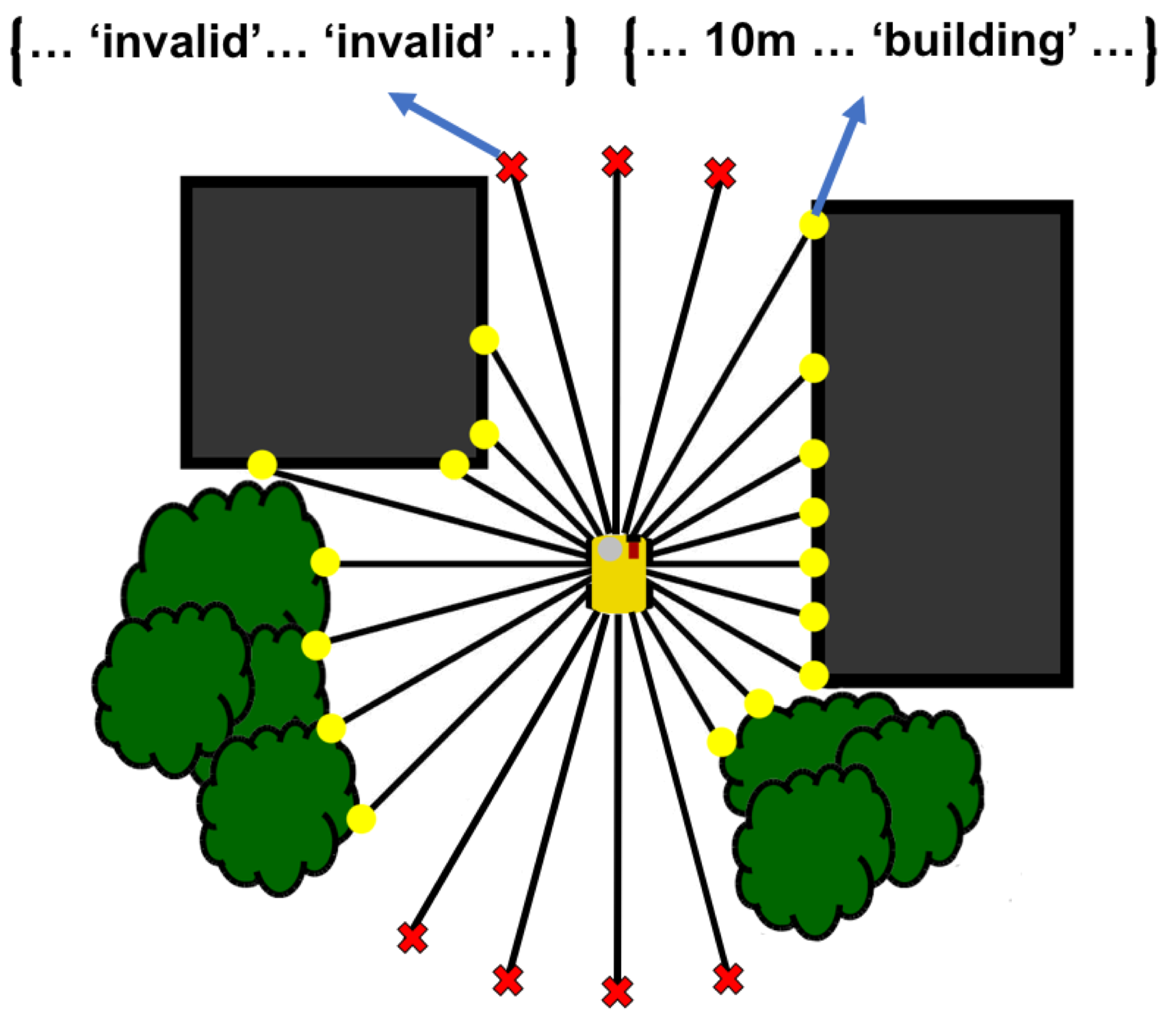}
	\caption{Illustration of the range and semantic descriptors. These descriptors are generated for both the \odem and the UGV data. Scan lines are generated at equally spaced angles ($\alpha$) up until a max distance. 
	If an obstacle is detected on a scan line, the distance to the obstacle and its semantic label are recorded at the appropriate elements. If no obstacle is detected, then invalid labels are recorded for the appropriate range and semantic elements. 
	}
	\label{fig:descriptors_illustration}
\end{figure}

To incorporate the semantic information into the descriptor, we assign the appropriate element of the descriptor to the semantic label of the segmentation at the location of where the obstacle was detected. If no obstacle was detected, then this element is also set to an `invalid' label.
An illustration of these descriptors is shown in \figref{fig:descriptors_illustration}. 

\noindent \textbf{Aerial Descriptors.} Before the UGV mission, descriptors are calculated for all $\numtraversable$ traversable pixels in the \odem.  We define the matrix of aerial descriptors for all traversable points as $\aerialdescriptors = \left\{\textbf{F}_{depth}, \textbf{F}_{semantic}\right\}$, where $\textbf{F}_{depth}$ and $\textbf{F}_{semantic}$ are matrices of size $\numscan \times \numtraversable$ that hold the range and semantic portion of the descriptors, respectively.

\begin{figure*}[ht!]
	\centering
	\begin{subfigure}[b]{0.48\textwidth}
		\centering
		\includegraphics[width=0.9\textwidth]{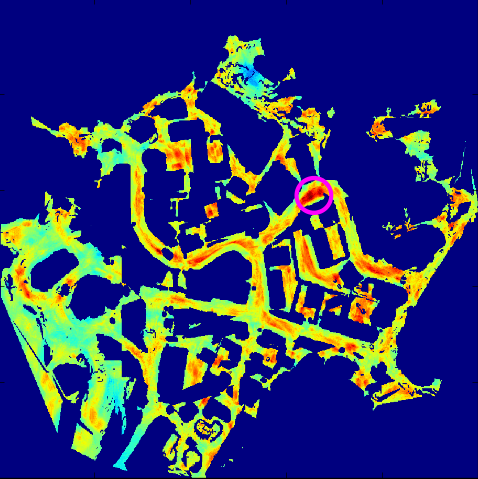}
		\caption{Heat map in an ambiguous region.}
		\label{fig:ambiguous_scores}
	\end{subfigure}
	\begin{subfigure}[b]{0.48\textwidth}
		\centering
		\includegraphics[width=0.9\textwidth]{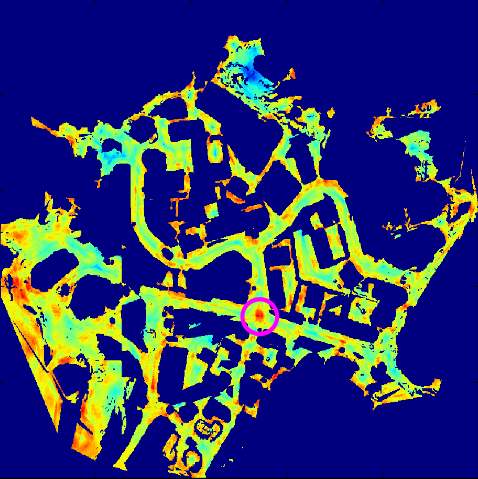}
		\caption{Heat map at an intersection.}
		\label{fig:intersection_scores}
	\end{subfigure}
	\caption{\gordonrb{Descriptor similarity heat maps, where the pink circles in each figure contain the ground truth location of the UGV. (a) Ambiguous region in between two buildings, where the UGV is at the street center. (b) UGV at an intersection (less ambiguous).}}
	\label{fig:scores_mats}
\end{figure*}

\noindent \textbf{Ground Descriptors.} We define descriptors of the UGV data for a particular time ($t$) and viewing angle in the aerial map ($\omega$) as $\ugvdesc(t, \omega) = \{\lidarfeat(t, \omega),\semanticfeat(t, \omega)\}$, where $\lidarfeat$ and $\semanticfeat$ are \gordonrb{$\numscan$-vectors} that hold the range and semantic portion of the descriptors. \gordonrb{Because of the UGV camera's limited field of view, we include} the term $\omega$ to define the view angle of the UGV. Changes to $\omega$ are represented by circular shifts of \gordonrb{$\aerialdescriptors$}.

In \secref{sec:experiments}, we show how we identify obstacles. By projecting the 3D points classified as obstacles into the $xy$ plane, we generate a 2D obstacle map similar to the one generated from the aerial data. We then feed this obstacle map for the laser data into the same function that generates the aerial descriptors. However, since we do not use an omnidirectional camera, it is only possible to obtain valid semantic labels for a subset of the scan lines.
During the descriptor similarity scoring process, we search over all values of $\omega$ to score each possible position $\aerialcoord$.

\noindent \textbf{Descriptor Similarity Scoring.}
Given a descriptor for the UGV data $\ugvdesc(t, \omega)$ and the set of aerial descriptors $\aerialdescriptors$, we search for the closest $\uavdesc \in \aerialdescriptors$ with a custom similarity measure. Binary vectors represent the element-wise similarity between $\uavdesc$ and $\ugvdesc(t, \omega)$, where we define elements of the range and semantic portions as

\begin{align}
    \delta_{range,i} &=
		\begin{cases}
		1, & \text{if}\ |\lidarfeati(t, \omega) - \textbf{F}_{depth, ji}| < \diffth \\
		0, & \text{otherwise}
		\end{cases}
	\\
	\delta_{semantic,i} &=
		\begin{cases}
		1, & \text{if}\ \semanticfeati(t, \omega) = \textbf{F}_{semantic, ji} \\
		0, & \text{otherwise}
		\end{cases}
\end{align}
where $\diffth$ is the max difference in distance allowed between the two descriptors being scored. We set $\diffth$ to 10\% of the max LiDAR distance (40m) on each scan line (4m) to accommodate differences in scale of the depth data and the different perspectives.
We do not set $\delta_{semantic,i}$ to 1 if one or both of the descriptor values of the segmentation are invalid (\ie no obstacle was found on the corresponding scan line of that element). 

We calculate the similarity between the descriptors as

\begin{equation}
s(\ugvdesc(t, \omega), \uavdesc) = \sum_i^n \delta_{range,i} + \gamma \delta_{semantic,i}
\end{equation}
where $\gamma$ is used to scale the segmentation score. 
There are a fixed number of scan lines that \emph{can} have valid semantic labels. However, not all scan lines in this subset will find obstacles, and therefore there are a variable number of valid semantic labels. This is the reason we scale the semantic score with $\gamma$.
\gordonrc{Given $\numscan$, the length of $\semanticfeat$,} and the number of valid labels for the segmentation portion of the current UGV descriptor ($v$), we \gordonrb{set} $\gamma = \frac{\numscan}{v}$.

We chose this similarity measure over alternatives (\eg Euclidean distance), because it is robust to small structural changes in the scene. 
\gordonrb{
For example, if a new obstacle appears in the scene after the UAV flight, then a subset of each UGV descriptor that observes the new obstacle will be affected. The aerial descriptors for the corresponding points in the \odem will not observe this obstacle, and will therefore potentially observe other obstacles much farther away. If Euclidean distance is used to measure similarity between the aerial and ground descriptors, then it is not likely these descriptors will match well. With our similarity measure, we can still score these descriptors as being similar, as long as the rest of the scan lines, that do not observe the new obstacle, match well.
}

We score each position of the UGV independent of previous predictions and odometry as

\begin{equation}
\textbf{S}(p) = \max_{\omega} \eqspacing s(\ugvdesc(t, \omega), \uavdesc),
\end{equation}
where $\uavdesc$ is the descriptor in $\aerialdescriptors$ that corresponds to position $p$.
Independent predictions for the UGV's location are made by finding

\begin{equation}
\hat{x}_t = \max_p \eqspacing \textbf{S}(p).
\label{eqn:position_estimates}
\end{equation}

By mapping these descriptor similarities to colors, we can display heat maps for each descriptor generated with the UGV's data. Two examples of this are shown in \figref{fig:scores_mats}. In \figref{fig:ambiguous_scores}, we show how positions at street centers with buildings on both sides can cause ambiguity. In \figref{fig:intersection_scores}, we show how more unique locations are easier to identify.

\subsection{Incorporating Odometry}
\label{sec:odom}

We obtain trajectories in GPS-denied environments for the UGV using \gordonrb{a GPS-free version of the SLAM approach described in \cite{gregory2016application}, where trajectories are generated using images, LiDAR, and the UGV's odometry.}
\gordonra{
Our problem can be viewed as mapping these trajectories to positions in the \odem. Our approach involves reasoning about local likelihoods from descriptor similarities and a prior generated using the UGV's estimated current position from the trajectory data. To calculate this prior,
}
we find a transformation using the position estimates ($\hat{x}_0 \ldots \hat{x}_{t-1}$) and the global trajectory ($x_0 \ldots x_{t-1}$), which has not been georegistered. We find transformations between this trajectory and our position estimates using RANSAC. At each iteration, we sample points and find Procrustes transformations. 
We define the predicted position from the current transformation as $\tilde{x}_t$, \gordonrb{around which we center the prior.}

Note that scale ambiguity is not a concern here, since we are mapping the trajectories from a local coordinate frame to the orthophoto. Therefore, even monocular SLAM approaches, such as \cite{mur2015orb}, can be used with our approach.

The reason we use previous independent predictions of position ($\hat{x}_0 \ldots \hat{x}_{t}$) at each iteration with RANSAC is that we do not want the next position estimate ($\tilde{x}_t$) to be too heavily influenced by the last estimate ($\tilde{x}_{t-1}$). For example, if the UGV is driving through an ambiguous part of the map, then it may not localize well until it reaches a more distinctive part of the scene. Incorrect predictions in ambiguous parts of the map tend to be scattered, and therefore removed as outliers when finding transformations with RANSAC.

\subsection{Particle Filter}
\label{sec:particle_filter}

We use a particle filter to exploit temporal information and predict the location of the robot at each iteration. Each particle calculates its weight using the likelihoods output by our descriptor similarities and the prior distribution of the UGV's next position. We define the position of the $i$th particle at time $t$ as $\particle$. Our likelihoods from our descriptor similarities are \gordonrb{defined as 
$\frac{ \descriptorscores(\particle)}{\sum_p \eqspacing \descriptorscores(p)}$,}
where $\descriptorscores(\particle)$ is the score in $\descriptorscores$ for $\particle$.
We use a prior distribution of the UGV's position for each particle \gordonrb{by sampling from $\mathcal{N}(\tilde{x}_{t}, \sigma ^ 2 \Sigma)$.}
At each time step we update the particles as

\begin{equation}
y^i_t = y^i_{t-1} + \tilde{x}_t - \tilde{x}_{t-1} + u_i, \quad u_i \sim \mathcal{U}[0,\lambda]
\end{equation}
where $\tilde{x}_t - \tilde{x}_{t-1}$ shifts the particles in the direction of the next estimated position $\tilde{x}_t$, and $u_i$ is a sample from a uniform distribution, where we set $\lambda = 15$ pixels (5.1m). 
The importance weight of each particle ($\importanceweight$) is then calculated as

\begin{equation}
\gordonrc{\importanceweight \propto \frac{ \descriptorscores(\particle)}{\sum_p \eqspacing \descriptorscores(p)} n_i, \quad n_i \sim \mathcal{N}(\tilde{x}_{t}, \sigma ^ 2 \Sigma)}
\end{equation}
where we normalize these importance weights and resample the particles with them. \gordonrc{We estimate the position of the UGV at each time step as}

\begin{equation}
\gordonrc{\bar{x}_t = \sum_i \importanceweight \particle}
\end{equation}
\gordonrc{where $\bar{x}_t$ is the weighted average of the particles' positions.}


\section{Experiments}
\label{sec:experiments}

\subsection{Systems and Setup}
\label{sec:experimental_setup}

We perform experiments at an urban test environment. The scene contains buildings, vegetation, grass, and roads. 
To capture the aerial imagery used to generate \gordonrb{our} \odem, we mount a GoPro camera to a 3DR IRIS.
The \odem is generated with Pix4D\footnote{https://pix4d.com/},
but other 3D reconstruction techniques~\cite{mur2015orb} may be used.

\begin{figure*}[ht]

	\begin{subfigure}[b]{0.325\textwidth}
		\centering
		\includegraphics[width=0.87\textwidth]{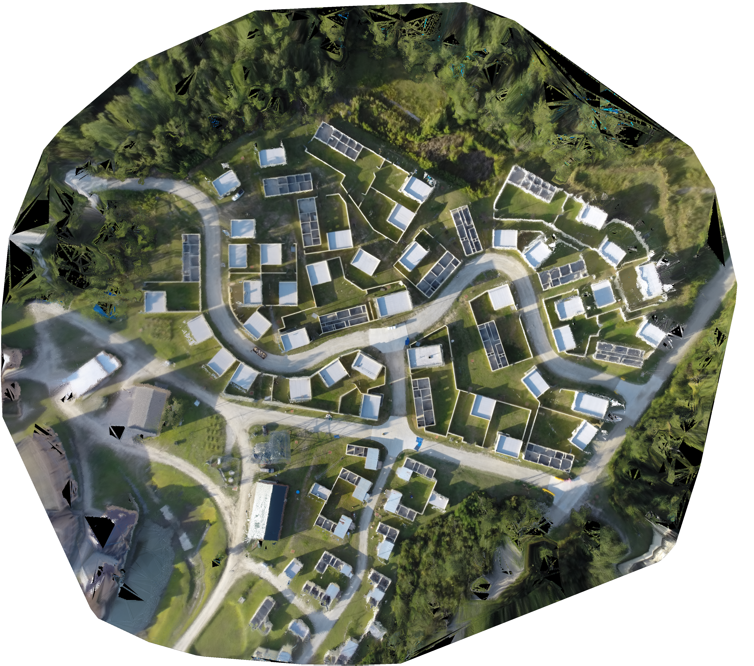}
		\caption{Orthophoto}
		\label{fig:orthophoto}
	\end{subfigure}
	\begin{subfigure}[b]{0.325\textwidth}
		\centering
		\includegraphics[width=0.87\textwidth]{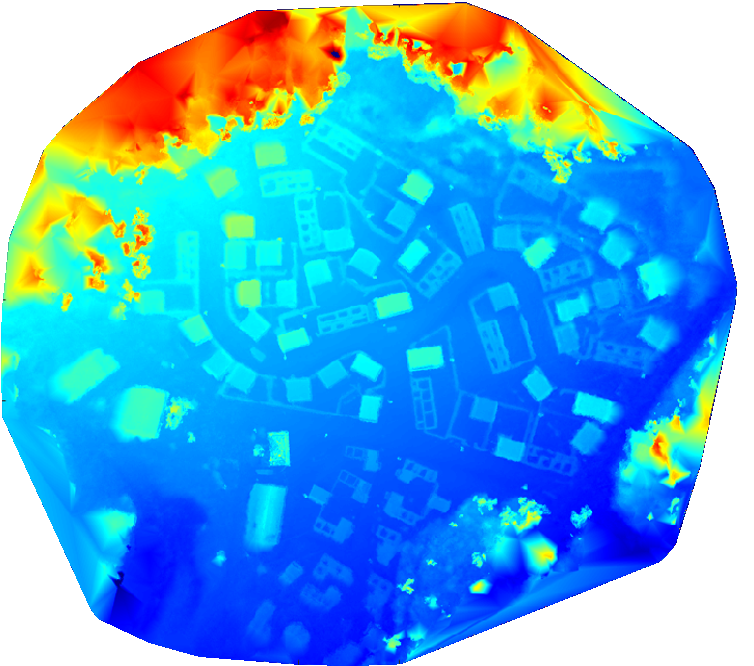}
		\caption{DEM}
		\label{fig:dem}
	\end{subfigure}
	\begin{subfigure}[b]{0.325\textwidth}
		\centering
        \includegraphics[height=2.9mm]{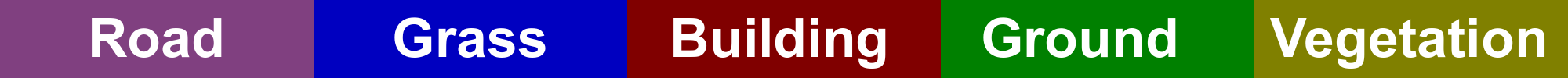}

		\vspace{0.3mm}

		\includegraphics[width=0.87\textwidth]{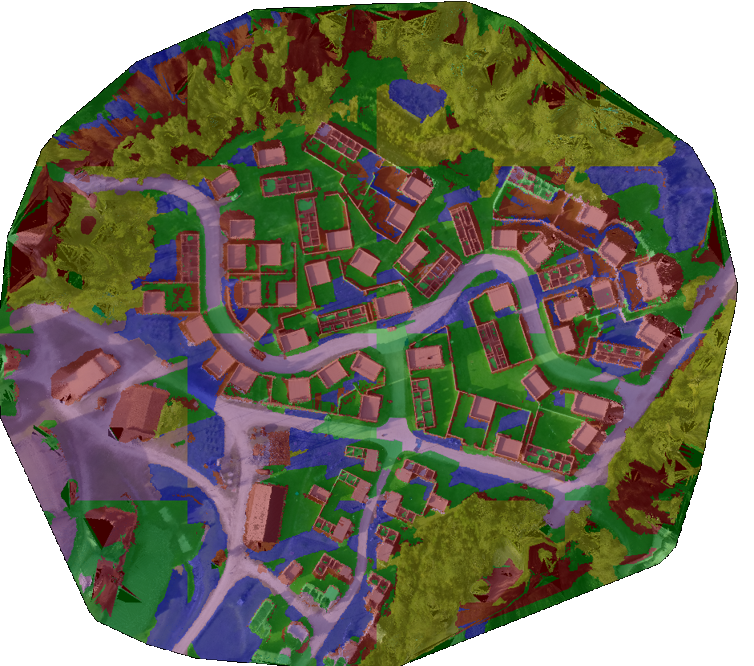}
		\caption{Segmentation}
		\label{fig:segmentation}
	\end{subfigure}
	\caption{The aerial data used in our experiments, which was generated using color images taken at an urban test site by a small, low-flying UAV. (a) is the orthophoto of the test site, (b) is the DEM, and (c) is the semantic segmentation generated using the orthophoto and DEM. The legend at the top shows colors of the semantic categories.}
	\label{fig:orthophoto_seg}
\end{figure*}

On the ground, color images and laser scans are captured by a Prosilica GT2750C camera and Velodyne HDL-32E LiDAR device, respectively. These sensors are calibrated and mounted on-board a Husky robot manufactured by Clearpath \cite{clearpath}. We use the calibration of the two sensors to project 3D points into the 2D images. By performing semantic segmentation on the 2D images, we can obtain semantic labels for a subset of the points in each laser scan.
We perform semantic segmentation on the \odem and the UGV's imagery by training on annotated aerial and ground image datasets, respectively, which contain no images at or near the test site. Range-based descriptors are computed using the digital elevation map (DEM) from the aerial data and the laser scans from the ground. 
We log GPS from the UGV and georegister the orthophoto only to obtain an approximate measurement of accuracy for our estimated position outputs. No geospatial information was used to estimate the location of the UGV in the \odem.

\subsection{Segmentation}
\label{sec:segmentation}

To perform localization without relying on low-level appearance-based features (\eg color, texture), we use segmentations of the \odem generated from the UAV imagery, and segmentations of the ground imagery and LiDAR. Below we describe each segmentation process.

\noindent \textbf{Aerial-view Segmentation}.
We use a two-stage approach to segmenting the \odem. First, we use the Automatic Labeling Environment (ALE)~\cite{ladicky2011thesis} to train a segmentation model using a dataset of 78 images, annotated with ground truth categories, captured from low-flying UAV in different environments. The orthophoto is then segmented using the trained model. The second stage of the process is to use the DEM to refine the segmentation. Using obstacles identified with the DEM, we correct the predictions of pixels inside of obstacle regions classified as traversable categories (\eg road) by reclassifying them as non-traversable categories (\eg building). We follow a similar procedure for pixels classified as non-traversable categories inside of non-obstacle regions.
We assume that roads will be confused with buildings, and that grass will be confused with vegetation. We use this rule-based approach to make the corrections. 

To identify obstacles in the DEM, we first generate an edge map. We initialize the set of ground pixels at the position farthest from an edge. The set of ground pixels is then iteratively expanded, where neighboring pixels with an elevation difference below a threshold 
are added to the set. Pixels that do not belong to the ground region are considered obstacles. The orthophoto, DEM and segmentation are shown in \figref{fig:orthophoto_seg}. The categories shown at the top of \figref{fig:segmentation} are the final set of categories for the segmentation. However, the segmentation model is trained with the category `shadow'. Traversable regions in the output classified as `shadow' are assigned the label `ground', and obstacle regions classified `shadow' are assigned the label `building'.

\noindent \textbf{Ground-view Segmentation}.
We segment images from the UGV's camera using the ALE~\cite{ladicky2011thesis} to train a model on a set of 100 annotated images taken from scenes outside of the test site. In a similar procedure to the segmentation of the aerial data, we identify obstacles in the laser scans to refine the semantic predictions of the 3D points. We represent the 3D point clouds as $\{x,y,h\}$, where $h$ is the height of the points. We downsample the points by rounding the $x,y$ values to the nearest tenth and then keep the unique points. 
Delaunay triangulation is run on the unique points $\{x,y\}$ to create an adjacency matrix for the points after projection to the $xy$ plane. For a pair of neighboring points, say $\{\left(x_{i},y_{i}\right), \left(x_{j},y_{j}\right)\}$, point $\left(x_{i},y_{i}\right)$ is said to be an obstacle if $h_{i} - h_{j}$ \gordonra{is greater than 0.2m.}
We use the same rule-based approach as the aerial segmentation to correct the confusion between roads vs. buildings, and grass vs. vegetation.

We make an assumption that we can be located on either road or grass. By predicting which surface type we are currently on, we can use that information to better localize the UGV by possibly reducing ambiguity. To predict the surface type, we simply use the image segmentation output by the ALE, and take the mode of the segmentation outputs for the bottom portion of the image. If we predict that we are on a road, then we set the scores for all grass regions to 0. Similarly, we set scores for road regions to 0 when we predict that the UGV is on grass.

\subsection{Results}
\label{sec:results}

When \gordonrc{comparing to GPS}, we consider two approaches: (1) We output our estimate of the position using only the information we have up until that point. This measures the accuracy of how well the UGV was able to localize itself in real-time. This is important for the UGV to make decisions immediately. (2) Given the entire trajectory at the end of the mission, we map it to the orthophoto. This measures how accurately the UGV was able to localize its position history, which may be important for subsequent missions.

\begin{figure}[ht!]
	\centering
	\includegraphics[width=0.92\columnwidth]{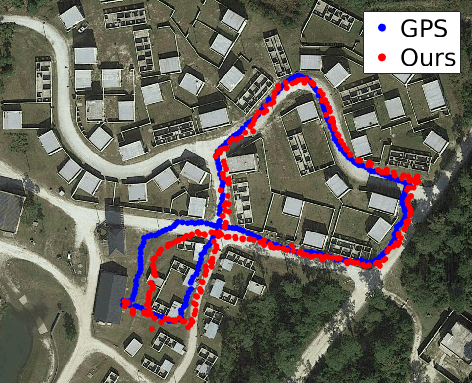}
	\caption{Google Maps overlay of path from GPS (blue) and our predictions (no GPS used). 
}
	\label{fig:google_maps_overlay}
\end{figure}

We present results in Table~\ref{tab:results}. 
\rangeonly and \rangeonlyfull only use the range portion of descriptors. 
The 
\rangeonly and \rangesemantic approaches predict
predict the position estimates with the data available up until the time of the current prediction. This is more challenging and tests the ability of the approaches to localize the UGV quickly so that real-time decisions can be made. 
The 
\rangeonlyfull and \rangesemanticfull approaches
predict the position estimates once the entire trajectory has been generated. This shows that improved localization can be performed once the whole trajectory becomes available.
To obtain these \gordonrc{results} we calculated the average \gordonrc{differences to GPS} of the full pipeline run for 20 iterations, where at each iteration we calculated the average \gordonrc{difference to GPS} over all positions. This was done because of the random element of our approach from RANSAC and the particle filter.

\begin{table}[h]
\centering
{\small
\scalebox{\tableScale}{
\begin{tabular}{ccccc}
\toprule  
\rangeonly & \rangeonlyfulltable & \rangesemantictable & \rangesemanticfulltable  \\
\midrule
5.392 $\pm$ 0.125 & 5.185 $\pm$ 0.285 & 4.676 $\pm$ 0.014 & \textbf{4.61} $\pm$ 0.088\\
\bottomrule
\end{tabular}
}
}
\caption{\gordonrc{Average difference to GPS} (meters) with standard errors for each approach (lower is better).}
\label{tab:results}
\end{table}

We show an overlay of our outputs for one of the runs with \rangesemantic and the GPS measurements on a satellite image of the test site in \figref{fig:google_maps_overlay}.
\gordonra{
This helps illustrate that GPS is not ground truth, since we know the UGV was located on the road, not entering and exiting the buildings. We also observe that the largest \gordonrc{difference with GPS} seems to occur when the UGV is in the grass near the left-most points of the path, near the end of the mission. We believe this \gordonrc{difference} is caused by a mixture of ambiguous descriptors and possible drift associated with the SLAM algorithm.
}
\gordonra{
We also observe that \rangesemantic generates less ambiguous descriptors in some regions than the range-only approaches. One example is near the right-most points of the path where the UGV is in between the trees and the buildings. Our approach correctly segments vegetation on one side of the UGV and buildings on the other side, which are incorporated into the descriptors in that area.
}

\section{Conclusions and Future Work}
\label{sec:conclusions}

We have demonstrated a successful approach for localizing a UGV in GPS-denied environments. The difference of our approach and GPS is under 5m for the complicated urban environment that we test at. The UGV uses images and laser scans to localize itself in a 2.5D orthophoto generated from aerial imagery captured by a low-flying UAV
\gordonrb{\emph{and} we show semantics help}. 
We represent the appearance portion of the aerial and ground data with semantic segmentations so descriptor similarity scores are robust to appearance-based scene changes. Our similarity measure to score pairs of descriptors also allows for small structural scene changes. 

One potential direction for future work would be to develop an active search strategy to assist in localization. The UGV could navigate to areas it believes will be less ambiguous to the localization algorithm. Another possibility would be to register UAV and UGV satellite imagery, where the systems could collaborate to perform localization. For example, the UAV could be used to quickly gather higher resolution imagery and depth data in areas that will potentially assist the UGV's ability to localize itself.

\section*{Acknowledgements}
This work was supported by the U.S. Army Research Laboratory through the Center for Unmanned Aircraft Systems.

{
\small
\bibliographystyle{IEEEtran}
\bibliography{gordon}

\begin{thebibliography}{10}
\providecommand{\url}[1]{#1}
\csname url@rmstyle\endcsname
\providecommand{\newblock}{\relax}
\providecommand{\bibinfo}[2]{#2}
\providecommand\BIBentrySTDinterwordspacing{\spaceskip=0pt\relax}
\providecommand\BIBentryALTinterwordstretchfactor{4}
\providecommand\BIBentryALTinterwordspacing{\spaceskip=\fontdimen2\font plus
\BIBentryALTinterwordstretchfactor\fontdimen3\font minus
  \fontdimen4\font\relax}
\providecommand\BIBforeignlanguage[2]{{%
\expandafter\ifx\csname l@#1\endcsname\relax
\typeout{** WARNING: IEEEtran.bst: No hyphenation pattern has been}%
\typeout{** loaded for the language `#1'. Using the pattern for}%
\typeout{** the default language instead.}%
\else
\language=\csname l@#1\endcsname
\fi
#2}}

\bibitem{hauff2012geo}
C.~Hauff and G.-J. Houben, ``Geo-location estimation of flickr images: social
  web based enrichment,'' in \emph{European Conference on Information
  Retrieval}.\hskip 1em plus 0.5em minus 0.4em\relax Springer, 2012, pp.
  85--96.

\bibitem{serdyukov2009placing}
P.~Serdyukov, V.~Murdock, and R.~Van~Zwol, ``Placing flickr photos on a map,''
  in \emph{Special Interest Group on Information Retrieval}.\hskip 1em plus
  0.5em minus 0.4em\relax ACM, 2009, pp. 484--491.

\bibitem{van2010combining}
O.~Van~Laere, S.~Schockaert, and B.~Dhoedt, ``Combining multi-resolution
  evidence for georeferencing flickr images,'' in \emph{International
  Conference on Scalable Uncertainty Management}.\hskip 1em plus 0.5em minus
  0.4em\relax Springer, 2010, pp. 347--360.

\bibitem{workman2015wide}
S.~Workman, R.~Souvenir, and N.~Jacobs, ``Wide-area image geolocalization with
  aerial reference imagery,'' in \emph{International Conference on Computer
  Vision}.\hskip 1em plus 0.5em minus 0.4em\relax IEEE, 2015, pp. 3961--3969.

\bibitem{lee2015predicting}
S.~Lee, H.~Zhang, and D.~J. Crandall, ``Predicting geo-informative attributes
  in large-scale image collections using convolutional neural networks,'' in
  \emph{Winter Conference on Applications of Computer Vision}.\hskip 1em plus
  0.5em minus 0.4em\relax IEEE, 2015, pp. 550--557.

\bibitem{weyand2016planet}
T.~Weyand, I.~Kostrikov, and J.~Philbin, ``Planet-photo geolocation with
  convolutional neural networks,'' \emph{arXiv preprint arXiv:1602.05314},
  2016.

\bibitem{kroeger2014video}
T.~Kroeger and L.~Van~Gool, ``Video registration to sfm models,'' in
  \emph{European Conference on Computer Vision}.\hskip 1em plus 0.5em minus
  0.4em\relax Springer, 2014, pp. 1--16.

\bibitem{david2011orientation}
P.~David and S.~Ho, ``Orientation descriptors for localization in urban
  environments,'' in \emph{International Conference on Intelligent Robots and
  Systems}.\hskip 1em plus 0.5em minus 0.4em\relax IEEE, 2011, pp. 494--501.

\bibitem{brubaker2013lost}
M.~Brubaker, A.~Geiger, and R.~Urtasun, ``Lost! leveraging the crowd for
  probabilistic visual self-localization,'' in \emph{Conference on Computer
  Vision and Pattern Recognition}.\hskip 1em plus 0.5em minus 0.4em\relax IEEE,
  2013, pp. 3057--3064.

\bibitem{brubaker2016map}
M.~A. Brubaker, A.~Geiger, and R.~Urtasun, ``Map-based probabilistic visual
  self-localization,'' \emph{Transactions on Pattern Analysis and Machine
  Intelligence}, vol.~38, no.~4, pp. 652--665, 2016.

\bibitem{viswanathan2014vision}
A.~Viswanathan, B.~R. Pires, and D.~Huber, ``Vision based robot localization by
  ground to satellite matching in gps-denied situations,'' in
  \emph{International Conference on Intelligent Robots and Systems}.\hskip 1em
  plus 0.5em minus 0.4em\relax IEEE, 2014, pp. 192--198.

\bibitem{gregory2016application}
J.~Gregory, J.~Fink, E.~Stump, J.~Twigg, J.~Rogers, D.~Baran, N.~Fung, and
  S.~Young, ``Application of multi-robot systems to disaster-relief scenarios
  with limited communication,'' \emph{Field and Service Robotics}, pp.
  639--653, 2016.

\bibitem{mur2015orb}
R.~Mur-Artal, J.~Montiel, and J.~D. Tard{\'o}s, ``Orb-slam: a versatile and
  accurate monocular slam system,'' \emph{Transactions on Robotics}, vol.~31,
  no.~5, pp. 1147--1163, 2015.

\bibitem{clearpath}
``Clearpath robotics husky,'' \url{http://www.clearpathrobotics.com/husky/},
  accessed: 2016-07-18.

\bibitem{ladicky2011thesis}
L.~Ladicky, ``Global structured models towards scene understanding,'' Ph.D.
  dissertation, Oxford Brookes University, 2011.

\end{thebibliography}
}

\end{document}